# A Novel Template-Based Learning Model

Mohammadreza Abolghasemi-Dahaghani [1,2], Farzad Didehvar[1], Alireza Nowroozi[1]

[1] Amirkbair University of Technology, Mathematics and Computer Science Department,
424 Hafez Ave, Tehran, Iran
[2] Institute for Research in Fundamental Sciences (IPM), School of Cognitive Sciences,
Niavaran Bldg., Niavaran Square, Tehran, Iran
mr.abolghasemi@gmail.com, {didehvar, ar_nowroozi}@aut.ac.ir

**Abstract.** This article presents a model which is capable of learning and abstracting new concepts based on comparing observations and finding the resemblance between the observations. In the model presented here, the new observations are compared with the templates which have been derived from the previous experiences. In the first stage, the objects are first represented through a geometric description which is used for finding the object boundaries and a descriptor which is inspired by the human visual system and then they are fed into the model. Next, the new observations are identified through comparing them with the previously-learned templates and are used for producing new templates. The comparisons are made based on measures like Euclidean or correlation distance. The new template is created by applying onion-pealing algorithm. The algorithm consecutively uses convex hulls which are made by the points representing the objects. If the new observation is remarkably similar to one of the observed categories, it is no longer utilized in creating a new template. In order to identify the new observations in each stage, all the previous observations and the learned templates are utilized. The existing templates are used to provide a description of the new observation. This description is provided in the templates space. Each template represents a dimension of the feature space. The degree of the resemblance each template bears to each object indicates the value associated with the object in that dimension of the templates space. In this way, the description of the new observation becomes more accurate and detailed as the time passes and the experiences increase. We have used this model for learning and recognizing the new polygons in the polygon space. Representing the polygons was made possible through employing a geometric method and a method inspired by human visual system. Various implementations of the model have been compared. The evaluation results of the model prove its efficiency in learning and deriving new templates.

**Keywords:** Concept Learning, Experience Representation, Templates, Polygon Recognition, Human Vision Nervous System